\documentclass[letterpaper,twocolumn,10pt, table,xcdraw]{article}
\usepackage{style/usenix2019_v3}
\usepackage{style/mist}

\usepackage{lipsum}
\usepackage{booktabs} 

\begin{document}

\title{\Large \bf Design and Implement an Enhanced Simulator for Autonomous Delivery Robot}

\author{Zhaofeng Tian}
\author{Weisong Shi}

\affil{Department of Computer Science, Wayne State University, Detroit, USA}  
\date{\today}

\maketitle

\begin{abstract}
As autonomous driving technology is getting more and more mature today, autonomous delivery companies like Starship, Marble, and Nuro has been making progress in the tests of their autonomous delivery robots. While simulations and simulators are very important for the final product landing of the autonomous delivery robots since the autonomous delivery robots need to navigate on the sidewalk, campus, and other urban scenarios, where the simulations can avoid real damage to pedestrians and properties in the real world caused by any algorithm failures and programming errors and thus accelerate the whole developing procedure and cut down the cost. In this case, this study proposes an open-source simulator based on our autonomous delivery robot ZebraT to accelerate the research on autonomous delivery. The simulator developing procedure is illustrated step by step. What is more, the applications on the simulator that we are working on are also introduced, which includes autonomous navigation in the simulated urban environment, cooperation between an autonomous vehicle and an autonomous delivery robot, and reinforcement learning practice on the task training in the simulator. We have published the proposed simulator in \href{https://github.com/Zhaofeng-Tian/ZebraT-Simulator.git}{Github}. 

\end{abstract}
\section{Introduction}
\label{sec:intro}

Autonomous driving technologies has been fast growing and continuously making progress, as a branch topic of autonomous driving, the autonomous delivery robot (ADR) landscape has been evolving rapidly. Domino proposed their ADR “DRU”, which is claimed to be the world’s first autonomous pizza delivery vehicle. Starship Technologies, a company founded in 2014, has announced that it will be rolling out its delivery robot services to corporate and academic campuses in the US and Europe\cite{Starship}. Marble, a startup company founded in April of 2017 and Dispatch, a San Francisco-based company has also been working on their own ADRs\cite{Marble}. Other companies like Nuro and Udelv that working on the on-road delivery are drawing more and more attention from the market \cite{nuro}\cite{udelv}.

Meanwhile, the concerns about the safety and potential impact that ADRs bring into human society are presented by some articles\cite{impact1}\cite{impact3}.  Some of the concerns are about the difference between Autonomous Vehicles (AVs) and ADRs. AVs drive on the main road with a well-structured map, e.g. a vector map, however, ADRs may need to drive on the sidewalk, closed campus, and residential area\cite{impact2}. To avoid potential risks in the aforementioned environments, e.g. hitting a pedestrian, crashing on a building, and losing its way due to sparse features fed to the sensors, simulations, and simulators are important. 

Thanks to Robot Operating System (ROS)\cite{ROS}, a framework for robot communication, and Gazebo, a simulation platform compatible with ROS\cite{gazebo}, developers can build simulation models with these tools. For example, Study \cite{sim1} \cite{sim2} both propose a simulation model for indoor race cars, and TurtleBot, a indoor robot created for research\cite{turtle}. Additionally, Gazebo model for vehicle  Lincoln MKZ, Ford F-150 and Ford Fusion are also proposed by an drive-by-wire company DataSpeed\cite{dataspeed}. 

Although the simulation method and simulation model are used in these studies, which focus more on autonomous driving vehicles or indoor robots, a simulator dedicated to ADRs and ADR's applications is rarely mentioned and studied. Since ADR's dimensional parameters and appearance are different from AVs and indoor robots, directly using AV or indoor robot models for ADR's research could cause inaccuracy because the motion planning algorithm could fail in certain ADR working conditions. Additionally, the simulation environment is also important in a simulator, the previous study only provides a model for the robot per se instead of testing the environment of dedicated applications. And as autonomous delivery is getting more interest from the market, there is a need to build a simulator consisting of simulation model and simulation environment for its related research to facilitate progress in the industry. To bridge the gap from the previous research to the simulation needs of the industry, this study proposes an enhanced simulator for the autonomous delivery robot and introduces three related applications we are working on.

In this study, ROS and Gazebo are leveraged to build a simulator for the ADR, where the simulator developing procedure from Computer-aided design (CAD) model to Gazebo model is illustrated in detail, which is rarely seen in other papers. Moreover, the simulation environment is also built for robot testing. And the simulator package will be open-source to the public for their own research needs. Besides, this study introduces three applications that can be working on our simulator. One of them is autonomous navigation in the simulated urban environment, where the robot and sensors are simulated in the simulator to sense the environment and plan a path as the real robot do in the real world. The second application is the cooperation between an autonomous vehicle and an ADR, testing the communication and data sharing functionalities. The third one is about Reinforcement Learning (RL),of which algorithms can be tested in the simulator to accelerate the training and promote the popularization of reinforcement learning methods in the real tasks.

Based on the aforementioned content, the contribution of this study to the society and researchers can be summarized as follows.
\begin{itemize}
  \item This study proposes a dedicated simulator consisting of robot model and simulation environment for the autonomous delivery robot and its applications, and it is open-source to all other researches to facilitate related studies, which can be considered as an innovation of this paper.
  \item This study details the procedure of building a robot simulator from CAD model to Gazebo model, which rarely seen in other research. 
  \item Besides simulator itself, we specifically introduce tree applications that we are working on the simulator, in which the practice of reinforcement learning is a quite new topic that is hardly studied by other articles.
\end{itemize}

The organization of the rest of this paper is arranged as follows. Section 2 provides a background for simulation study in the robotic field and its benefits to the current research. Section 3 illustrates the procedure of modeling ZebraT and simulator development. Then, the applications and feasible practice on the simulator are introduced case by case in Section 4, while conclusion is draw in Section 5.

\section{Background and Motivation}
\label{sec:motivation}

The ADR landscape has been evolving rapidly, thus there is a huge demand for product tests corresponding to different operating domains. For example, an ADV that runs on the sidewalk needs to be tested in an outdoor environment with a sidewalk and potential pedestrians. Also, an ADV that works in a restaurant needs to be tested in an indoor environment with tables and chairs. So, a question comes up, what if you do not have an appropriate site for the environment requirements? What if you do not have enough financial support for the testing tools and hiring testers?  What if the robot hits a pedestrian due to an algorithm failure on the sidewalk? The simulation then will be a good solution to the questions above.

Simulation can play an important role in the testing of delivery robots in a simulated urban, pedestrian environment, and thus accelerate the product landing and improve the safety of both robots and pedestrians. In the past decades,  simulations are broadly used and simulators became an essential part of the autonomous driving and robotics research field.

Study\cite{kaur} introduces multiple simulation software and frameworks for testing self-driving cars. Also, the comparisons among the simulation software are shown, which gives other researchers guidance on choosing an appropriate simulator for their own research.  Robot Operating System (ROS) is also mentioned in the study, which is a popular open-source framework that has a large number of communities and developers\cite{ROS}.

Thanks to plenty of libraries, packages, and interfaces with other software of frameworks, ROS accelerates researchers’ study and provides solutions to the related industries. Meanwhile, Gazebo simulation software is officially adopted to be the simulation platform for ROS\cite{gazebo}, where the developer can build up some physical models while using the communication functions from ROS without worrying about compatibility problems. 

Study\cite{sim1} simulates a indoor race robot in Gazebo environment, transferring the appearance of the robot into the simulation environment, to secure that model in the simulation environment looks as real as the robot in the physical world.  Additionally, study\cite{sim2} introduces how to simulate an Ackermann-steering-based indoor race car with ROS and Gazebo, and illustrates the communication mechanism between ROS nodes and the model in the Gazebo. TurtleBot, a famous indoor robot with its Gazebo model is widely used by researchers and developer to test algorithms in the indoor environments\cite{turtle}. Besides those indoor robot model, vehicles like Ford Fusion, F-150 and Lincoln MKZ are simulated by the campany DataSpeed in Gazebo\cite{dataspeed}. 

Besides, reinforcement learning method, a method enables AlphaGO to win the best human go player\cite{alpha}, still have not been popularized in other industries like automotive and robotics industry since it requires algorithm iterations to get the optimal policy that are not practical in a real car or robot. Thus, a simulator is particularly useful for algorithm iterations, and significantly cut down the total training time to get a good policy, which can be then transferred to a real car or robot.

From the studies mentioned above, we can highlight the benefits of using simulation methods in robotic research as follow.  

\begin{itemize}
  \item Simulations allow researchers to test their algorithms and concepts even if the required hardware is not available for them.
  \item Simulations contribute to the fast detection and correction of fatal errors in the algorithm or program. 
  \item The cost of doing simulations is comparatively lower than the physical experiments in the real world.

  \item Simulations avoid accidents caused by unknown risks that may be met in the real world and lead to hazards and damage. 
\end{itemize}

However, these studies only model the indoor robots or vehicle, of which size is too small or too big compared to ADR. If researches directly use these models to substitute the ADR model in a simulation task, the different size could impact the accuracy of path planning and motion planning. For example, if the ADR needs to work in a narrow alley, to use vehicle model could cause the path planning failure and to use indoor robot model could cause the different motion planning when encountering an obstacle. Thus, there is a need to build up an ADR simulation model for the ADR related research to enhance the simulation accuracy. Movever, since the simulation environment of ADR is different from AV's and indoor robot's. So the appropriate simulation environments are also important.

To bridge the research gap mentioned above, this study builds up a simulator an autonomous delivery robot with the robot model and simulation environments, and aims to share the simulator with the public to accelerate not only our future research but also other researchers’ research, bridging the gap between theoretical innovations and physical practice.

\section{Methodology}
\label{sec:methodlogy}

The methodology of developing a simulator is to extract the mathematical and physical features from the object in the real world, then formulate these features to the model in the simulator. For example, to develop a simulator for the ADR, we need to get the information of its body size, wheelbase, wheel radius, which are the important parameters of its kinematic model. What is more, the steering mechanism of the robot will be taken into the consideration, which determines the motion of the agent. Additionally, besides the mathematical model, the visualization of the agent is also important to give users a sense of reality. In this case, we use our ADR ZebraT as a model to design and implement this simulator, which can be generalized to other delivery robots for other researchers. 

Thus, in this section, we first introduce our ADR ZebraT about its dimension information and equipped sensors in subsection 3.1. We secondly illustrate how to import the 3D model of ZebraT into the simulator and define the transformation relationship among each link and joints in subsection 3.2. Then we build up the kinematic model for the agent in subsection 3.3, which is based on its Ackermann-steering mechanism. Simulation environments in the Gazebo is introduced in subsection 3.4.

\subsection{ZebraT introduction}
ZebraT is an electric-drive autonomous delivery robot equipped with a RoboSense 16-line Lidar sensor, an Intel D455 RGB-D camera, a processor with i5 8250U CPU, an Intel UHD Graphics 620 GPU, and an STM32 single-board chip for the communication between the processor and the motor drivers. Besides those key components, sensors like Inertia Measurement Unit (IMU), Global Navigation Satellite System (GNSS), and current measurement sensors are integrated, which are shown in Fig.1. Additionally, there are 2 storage cells on each side of ZebraT for the delivery purpose.  Also, the Ackermann steering mechanism driven by a 24V 400W BLDC motor makes ZebraT steer like a real car and have more load capacity. And two more same motors are used to drive the two real wheels to move the robot car. These components are also recorded in Table 1. 

In terms of software, Linux Ubuntu 16.04 LTS operating system with ROS Kinetic version is installed in the processor. Thanks to the integration of these hardware and software, ZebraT is capable of testing and validating algorithms for autonomous driving in both indoor and outdoor environments.

\begin{figure}[H] 
\centering 
\includegraphics[width=\columnwidth]{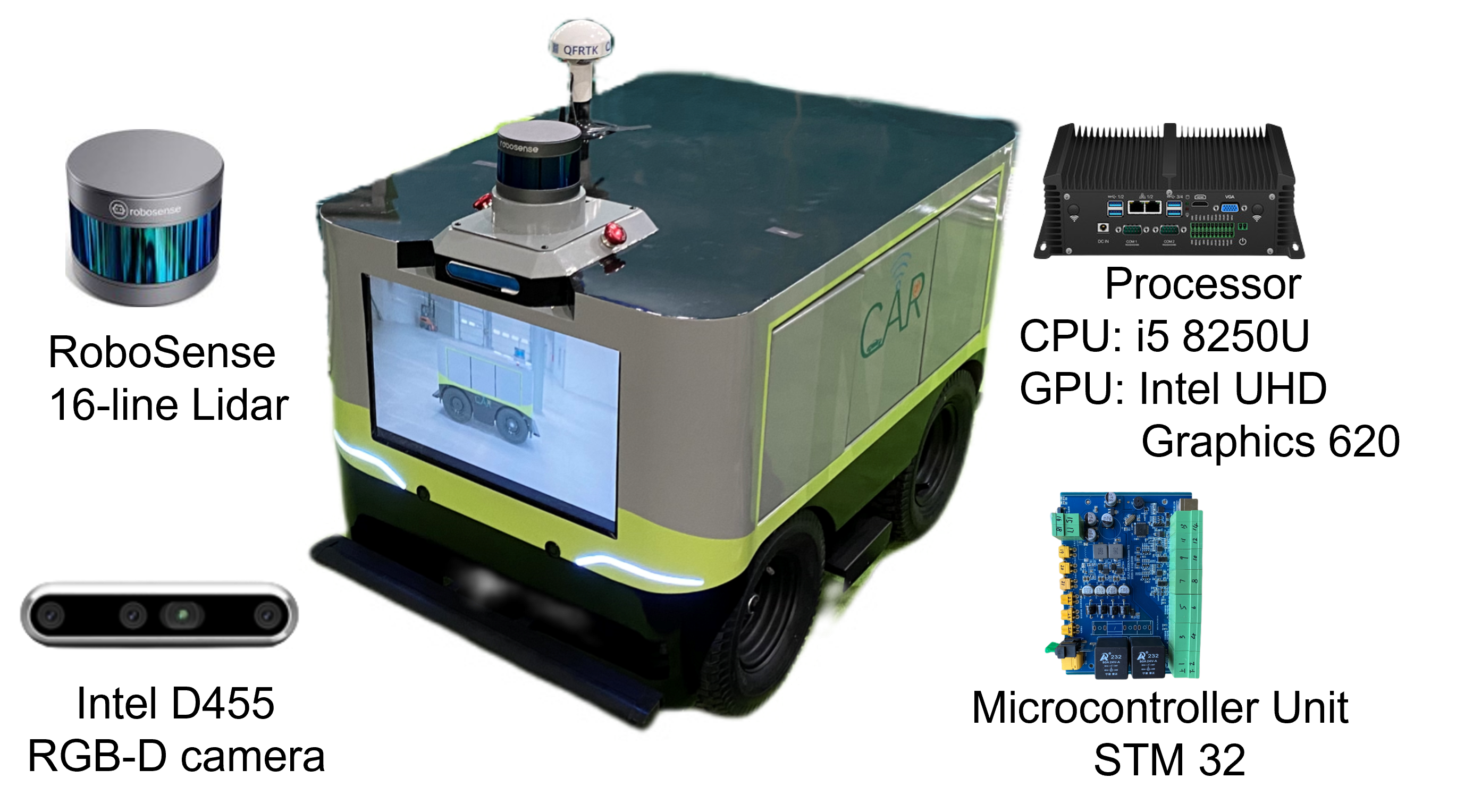} 
\caption{ZebraT} 
\label{Fig.main2} 
\end{figure}

\begin{table}[]

    \centering
    \begin{tabular}{c c c}
    \toprule
        Component &  Model & Quantity\\
    \midrule
        Lidar & RoboSense 16-line & 1 \\
        Camera & Intel D455 RGB-D & 1 \\
        CPU & i5 8250U & 1 \\
        GPU & Intel UHD Graphics 620 & 1 \\
        Microcontroller & STM32 & 1\\
        IMU & - & 1 \\
        GNSS & - & 1\\
        Current Sensor & - & 4 \\
        Steering Motor & 24V 400W BLDC & 1\\
        Driving Motor & 24V 400W BLDC & 2 \\
    \bottomrule
        
    \end{tabular}
    \caption{ZebraT Components}
    \label{tab:table1}
\end{table}
Besides the hardware and software introduced above, the other important factor of the robot modeling is the dimension information or so-called size.  Based on the dimension of the body shell, wheel diameter, robot length, width, and heights, we can calculate the coordinates of the centers of the body shell and wheels that the centers will be used to construct collision geometry for the collision detection in the later simulation, and also to calculate inertias using inertia formulations. Here we list all the dimension information of ZebraT in Table 2, which could be used in the simulator development. Note that body shell height is not equal to robot height since the wheel bottom is lower than the body shell bottom. 

\begin{table}[]
    \centering
    \begin{tabular}{c c}
    \toprule
       Dimension  &  Value(m)\\
    \midrule
        Wheel radius & 0.150\\
        body shell length & 0.963 \\
        body shell width & 0.672 \\
        body shell height & 0.557 \\
        total height & 0.640 \\
        wheel base & 0.530 \\
    \bottomrule
    \end{tabular}
    \caption{Dimension Information}
    \label{tab:dimension}
\end{table}

\subsection{CAD Model to Simulator Model}
Computer-Aided Design (CAD) technology is intensively used in the mechanical, electric, and robotic industries. Most design work nowadays is started from building a CAD model in the CAD software so that the structure validation and physical parameter calculation can be implemented earlier than the real product coming out. Our design also adopted this method, and the designed CAD model of ZebraT is shown in the top left of Fig. 2. 

While how to leverage the CAD model in the simulator is still a problem. Thanks to the ROS and Gazebo, which make this process easier.  As introduced in the previous section, Robot Operating System (ROS) is a popular open-source framework that provides a communication mechanism that applies to all kinds of robots including robot arm, differential-drive robot, Ackermann-steering robot, Unmanned Aerial Vehicle (UAV), and Autonomous Vehicle(AV). Additionally, Gazebo provides an open-source 3D simulator integrated with ROS that can simulate various types of robots mentioned above. Meanwhile, Gazebo provides users an opportunity to build their own robots using some fundamental elements like boxes, cylinders, and balls. More than that, Gazebo is able to simulate sensors like Lidar, camera, and IMU using customized plugins from the manufacturers. Gazebo supports robot models in SDF  and URDF formats, in which the robot links and joints are defined with coordinates, parental relationships, and motion types. Gazebo also allows importing meshed in the STL or DAE form to make the object model more realistic. 

In our study, we firstly mark the geometry centers of the body shell and wheels, and rotation axles of the steering mechanism and four wheels of ZebraT in the CAD model, then export their meshes separately. Also, we use the URDF file in the Gazebo Platform, which is an XML format file that describes the connection and transformation relationship between connected links and joints. All links and joints built up in the URDF model are respectively listed in Table 2 and Table 3.

\begin{table}[]
    \centering
    \begin{tabular}{c c}
    \toprule
       Link Name  &  Description\\
    \midrule
       \emph{base\_link} & Body shell\\
        \emph{lstr\_link} &  Left Steering Mechanism\\
        \emph{rstr\_link} & Right Steering Mechanism \\
       \emph{fl\_wheel} & Front Left Wheel \\
       \emph{fr\_wheel}& Front right Wheel \\
       \emph{rl\_wheel}& Rear Left Wheel \\
       \emph{rr\_wheel}& Rear Right Wheel \\
    \bottomrule
    \end{tabular}
    \caption{Links Description}
    \label{tab:links}
\end{table}

\begin{table}[]
    \centering
    \begin{tabular}{c c c}
    \toprule
       Joint Name  &  Connected Links & Joint Types\\
    \midrule
       \emph{base2lstr} & \emph{base\_link \& lstr\_link} & Revolute \\
       \emph{base2rstr} & \emph{base\_link \& rstr\_link} & Revolute \\
       \emph{fl\_axle} & \emph{lstr\_link \& fl\_wheel} & Continuous\\
       \emph{fr\_axle} & \emph{rstr\_link \& fr\_wheel} & Continuous\\
       \emph{rl\_axle} & \emph{base\_link \& rl\_wheel} & Continuous\\
       \emph{rr\_axle} & \emph{base\_link \& rr\_wheel} & Continuous\\

    \bottomrule
    \end{tabular}
    \caption{Joints Description}
    \label{tab:joints}
\end{table}
Note that there are three main types of joints we used in this study. They are, revolute, continuous, fixed joints, where revolute joint represents that the son link can rotate based on an axle, but has an angle constraint. In our case, the steering link cannot rotate beyond -60 to 60 degrees, so the base2lstr joint and base2rstr joint have a constraint for rotation as -60 to 60 degrees. The continuous joints, however, do not have angle limitations so that they can rotate continuously. Thus, our four wheels are all connected with continuous joints. Sensors like Lidar, camera and IMU are not necessary to rotate or move, so they are connected to the body shell with fixed joints.

What is more, for the purpose of simulating the sensors in ZebraT in the Gazebo environment as well, we add the sensor models in the URDF format for the sensor experiments, which are provided by the sensor manufacturers and ROS developers that we can include them in our main URDF model and then plugin them in Gazebo environment.  The usage of these sensors is shown in Table 4.
\begin{table}[]
    \centering
    \begin{tabular}{c c c}
    \toprule
       Sensor Type  &  URDF file & ROS Package\\
    \midrule
       Lidar &  VLP-16.urdf &\emph{velodyne\_description}\\
       Camera & camera.urdf & \emph{zebrat}\\
       IMU & imu.urdf & \emph{zebrat}\\

    \bottomrule
    \end{tabular}
    \caption{Sensors Description}
    \label{tab:sensors}
\end{table}

\begin{figure}[H] 
\centering 
\includegraphics[width=\columnwidth]{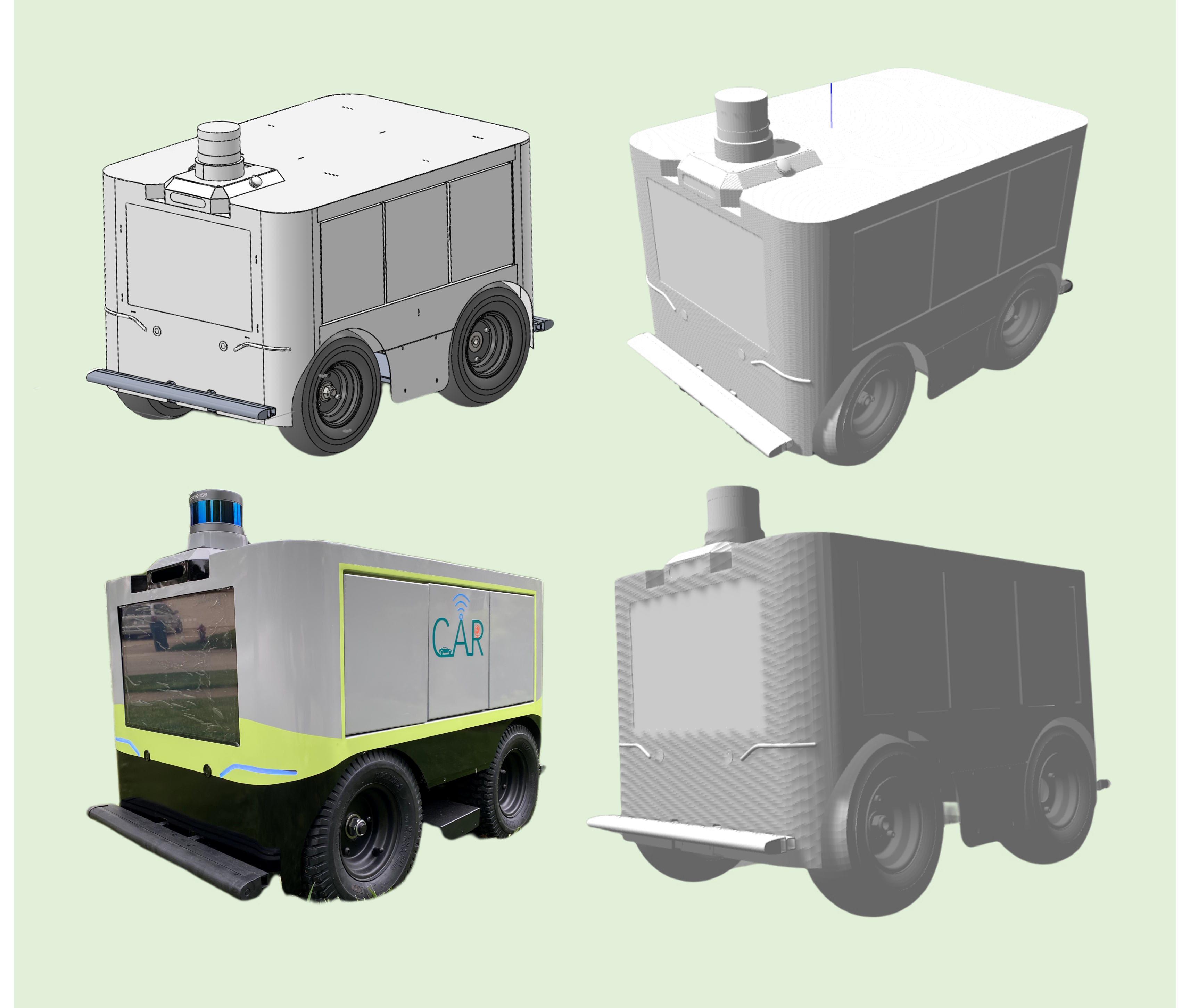} 
\caption{Results Comparison} 
\label{comparison} 
\end{figure}

After the links and joints are well defined, we need to calculate the inertia for each link in the simulation model because inertia is one of the most important physical attributes of an object not only in real world but also in the simulator, it can impact object's dynamic performance and stability in the initialization stage in the Gazebo environment. We found that use the inertia generated in CAD software for each link is not accurate, which caused model drifting and rolling in the Gazebo environment and even failures. To correct the faults, we consider the body shell of ZebraT as a box, the wheels as cylinders to simplify the inertia calculation and make the inertia accurate, so that the model can be stabilized in the initialization stage.

To calculate the inertia of a box for the body shell, we use the equation 1 - 3. Where \emph{I}$_{l}$,\emph{I}$_{w}$,and \emph{I}$_{d}$ represent the inertia respectively along the length, width and depth direction, \emph{l}, \emph{w}, \emph{d} represent the length, width and depth of the box, while \emph{m} represents mass.

\begin{equation}
I_{l}  = \left(\frac{m}{12}  \right)\left ( w^2 + h^2\right ) {\mathfrak{} \Large }     
\end{equation}
\begin{equation}
I_{w}  = \left(\frac{m}{12}  \right)\left ( l^2 + h^2\right ) {\mathfrak{} \Large }     
\end{equation}
\begin{equation}
I_{d}  = \left(\frac{m}{12}  \right)\left ( l^2 + w^2\right ) {\mathfrak{} \Large }     
\end{equation}
The inertia of wheels which are considered as cylinders, we use the equation 4 - 6. Where the \emph{I}$_{x}$,\emph{I}$_{y}$,and \emph{I}$_{z}$ represent the inertia respectively along the x, y and z direction referred to the world coordinate frame Gazebo environment, while \emph{r} represent the radius of the cylinder, \emph{h} denotes the height, and \emph{m} is the mass of the cylinder. 
\begin{equation}
    I_{x} = \frac{m}{12}\ast \left ( 3\ast r^2 +  h^2 \right ) 
\end{equation}

\begin{equation}
    I_{y} = I_{x}
\end{equation}

\begin{equation}
    I_{z} = \frac{m \ast r^2}{2} 
\end{equation}

After the inertia calculation, the model can be properly spawn into the Gazebo environment. The generated simulation model in Gazebo environment are shown in the right side of Fig. 2. Meanwhile, the CAD model is shown in the top left, and the real robot is shown in the bottom left. As we can see, the model in the simulation is quite realistic that can meet the normal requirements of most developers and users.

\subsection{Kinematic Model}
In our study, ZebraT is using the Ackermann-steering mechanism that is used in most cars. The attribute of this kind of steering mechanism is nonholonomic, which means it makes the agent cannot move along the lateral direction unless the agent’s speed on the longitudinal direction is not zero, imagine that your car cannot move to its side direction when it is still parked \cite{carlike}. So, the simulated agent, i.g. ZebraT in the simulator should be subject to the rule of this kind of kinematic constraints. The equation 7-9 formulate nonholonomic constraints for  the robot.
\begin{equation}
   \dot{x} = \cos \theta v
\end{equation}

\begin{equation}
   \dot{y} = \sin \theta v
\end{equation}

\begin{equation}
    \dot{\theta} =\frac{v}{L}\tan \phi   
\end{equation}

In the equations, $x$ and $y$ denote the location of the center of the axle between the two rear wheels with respect to the world coordinate frame in Gazebo environment. $\theta$ denotes the orientation of the car-like mobile robot with respect to the $x$ axis, $\phi$ is the steering angle with respect to the robot body and $\dot{y}$ represents the velocity of the rear wheels while $L$ is the wheelbase, which is shown as Fig. 3.  ($v$, $\phi$) are two control inputs to the kinematic model, which corresponds to the throttle and steering wheel input in the real car.

\begin{figure}[H] 
\centering 
\includegraphics[width=\columnwidth]{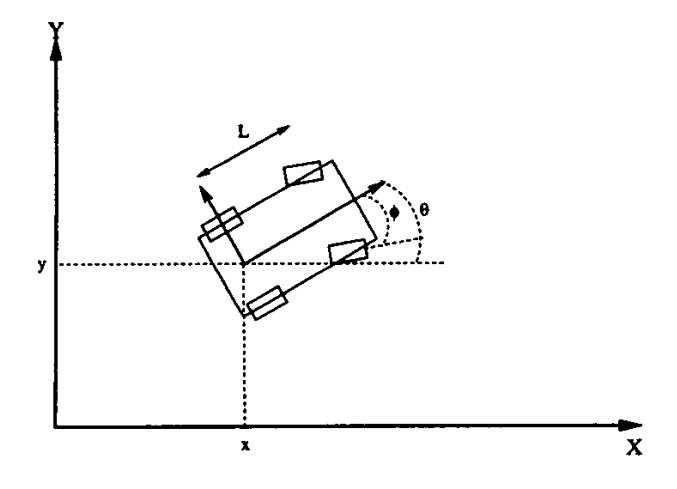} 
\caption{Kinematic Model} 
\label{kinematic model} 
\end{figure}
After formulating the kinematic model, we still need to transform the v. phi input into the velocity of six joints in the robot, since there are four wheels and two steer links that finally determine the motion of the robot. This kind of mapping relationship is described in Table. 6, where $joint\_angle$ denotes the rotation angle of the joint with the unit $rad$, $joint\_velocity$ denotes the rotation speed of the joint with the unit $rad/s$, and $r$ represents the radius of the wheel.

\begin{table}[]
    \centering
    \begin{tabular}{c c}
    \toprule
       Joint  &  Relationship to Input\\
    \midrule
       \emph{base2lstr} & $joint\_angle = \phi$\\
       \emph{base2rstr} & $joint\_angle = \phi$\\
       \emph{fl\_axle} & $joint\_velocity = v / r$\\
       \emph{fr\_axle} & $joint\_velocity = v / r$\\
       \emph{rl\_axle} & $joint\_velocity = v / r$\\
       \emph{rr\_axle} & $joint\_velocity = v / r$\\

    \bottomrule
    \end{tabular}
    \caption{Joint Control}
    \label{tab:joint control}
\end{table}

\subsection{Simulation Environment}
We select two specific simulation environment to test ADR. One is urban scenario, shown in Fig. 4. The other one is narrow alley, shown in Fig.6. The reason we select these two scenarios is that most ADRs need to work in the urban environment which consists of buildings, roads, and obstacles, and the simulated urban scenario is able to test the sensing, planning, capability of the robot in such complex environment. While the narrow alley could test the passability of the ADR in such narrow  environment where the traditional path planning methods could fail, and allow us to test new algorithm for this certain scenario. These two simulation environments are also listed in Table 7.

\begin{table}[]
    \centering
    \begin{tabular}{c c}
    \toprule
       Environment  &  Description\\
    \midrule
       \emph{Urban Scenario} & builds, roads, obstacles\\
       \emph{Narrow Alley} & walls to simulate alleys\\

    \bottomrule
    \end{tabular}
    \caption{Simulation Environments}
    \label{tab:env}
\end{table}

From above steps, the ADR simulator is designed and implemented where the robot model can be controlled by ROS node to implement the kinematic model that enables the simulated robot to move like the real one, and sensing and planning algorithms can be tested in the proposed simulation environments. However, specific applications of this simulator could be more important for users to work on their expected tasks. Thus, the applications that could leverage this simulator are introduced in the next section.

\section{Applications}
To accelerate the development of the product, the testing of algorithms, and the validation of the program, simulation is a good solution. The applications of the simulation method can vary from each industry. For instance, the traditional automotive industry may use simulation to test vehicle dynamics, driving comfort, and fuel economy, which is more related to the vehicle per se. While the autonomous driving industry is interested in sensing the environments with multi-type sensors, behavior planning, and motion planning \cite{kaur}. In our study, our simulator mainly focuses on autonomous driving-related applications like sensing, detection, and navigation. What is more, reinforcement learning, one of the hottest AI topics, can be also tested in our simulator, which is illustrated in detail later.
   
Although this study does not cover all the bases, we pick three representative applications in this section that we are studying now with this simulator, to provide a rough knowledge of autonomous driving and reinforcement learning. 

\subsection{Autonomous Navigation}
In the simulator, users and developers are able to build their customized environment, the so-called “world” in Gazebo platform. As Fig. 4 shows, the urban environment can be built as a “world” with buildings, roads, and some obstacles to test the ADV. Meanwhile, sensing technologies like mapping and localization can be implemented for autonomous navigation. We tested gMapping package in ROS\cite{gmapping} and other mapping algorithms e.g. Lego-LOAM\cite{lego}. 
\begin{figure}[H] 
\centering 
\includegraphics[width=\columnwidth]{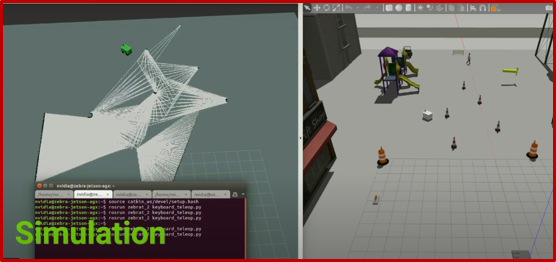} 
\caption{Navigation in the Simulator} 
\label{nvi} 
\end{figure}
Besides sensing technology, global path planning algorithms like Dijkstra and A star,  local path planning algorithms like Dynamic Window Approach(DWA)\cite{dwa}and Timed Elastic Band (TEB)\cite{teb} can be tested in the simulator. 

\subsection{Multiple Agents Cooperation}
Cooperation between Autonomous Vehicles (AVs) and Autonomous Delivery Robots (ADRs) is an interesting topic that AVs could first deliver goods to a community or campus, then ADRs could be responsible for the last mile delivery to the customers. Our simulator could test this process in terms of communication and planning between an ADR and an AV.  Fig. 5 shows our AV Hydro and ADR ZebraT, which can share the map and sensing data between each other to implement cooperation tasks. Although this is a new topic and not too much work has been done, the potential impact that it could bring to society still deserves more effort from developers and researchers.

\begin{figure}[H] 
\centering 
\includegraphics[width=\columnwidth]{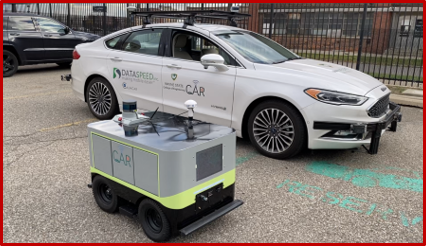} 
\caption{Cooperation between AV and ADR} 
\label{nvi} 
\end{figure}

\subsection{Reinforcement Learning}
Reinforcement Learning (RL) is a branch technology of machine learning. Unlike supervised learning which requires human supervision on data classification, RL makes agents learn from the practice from itself based on predefined reward functions, which are considered as the real intelligence for machines and robots. However, one important factor of RL has been limiting this technology intensively landed in the real world, that is the requirements for a huge number of iterations to improve the policy. Although RL has gained some achievements in some traditional games e.g. go given that AlphaGo has won the top human player\cite{alpha}, it has not been practiced too much on the agents like cars and robots due to the difficulties of intensive training of them in the real world.

Our simulator could be an appropriate solution for RL practice because the agent can be trained iteratively to accelerate the policy improvement, and thanks to the ROS framework, the improved policy can be transferred to the real robot so that the training time could be prominently cut down and the cost of real experiments could be exempted. 

We have done some experiments of RL task using our simulator and OpenAI\cite{openai}, a python library for RL training, which is shown in Fig. 6, where the robot needs to pass the alley. In this task, the robot could sense the environment via Lidar and then downsample the points data to generate states, based on which the RL algorithm can perform the iteration process.
\begin{figure}[H] 
\centering 
\includegraphics[width=\columnwidth]{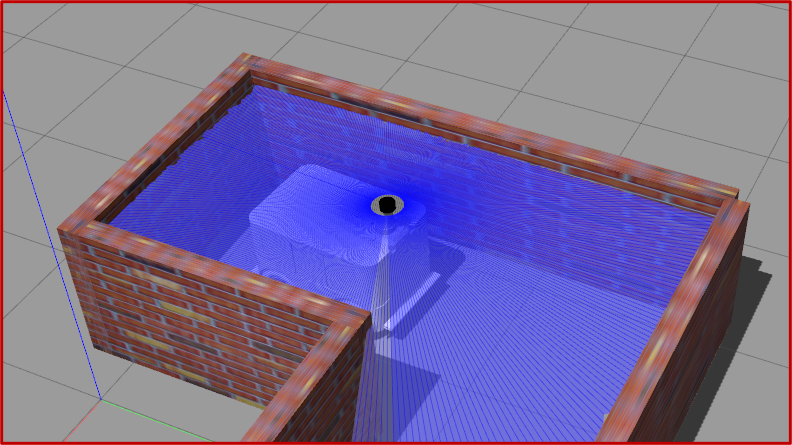} 
\caption{Reinforcement Learning Practice} 
\label{rl} 
\end{figure}
\section{Conclusion}
\label{sec:conclusion}
In summary, this study proposes an open-source simulator of our autonomous delivery robot ZebraT in ROS framework and Gazebo simulation platform. The specific procedure of developing this simulator is explained in section 2, which includes fundamental information about ZebraT, the developing process from CAD model to simulator model, and kinematic modeling of the Ackermann-steering mechanism, thanks to which the generated simulation model in the simulator not only looks realistic as the real one but also move like the real one. 
Furthermore, three applications we are studying with this simulator are introduced. The delivery robot can implement autonomous navigation in the simulated urban scenario with buildings, roads, and obstacles. Also, the data sharing between our AV Hydro and ADR ZebraT could be practiced in the simulator. Finally, the importance of our simulator to the popularization of reinforcement learning is explained with our own experience on the task training project using this simulator.
Of course, there are many other related topics that we have not covered in this paper, but we hope other researchers can make progress on them using our proposed simulator which will be published on Github.  And hopefully, we can bring more research to the public about the applications of this simulator and contribute to society.

\newpage
\bibliographystyle{plain}
\bibliography{reference/mist}
\end{document}